\documentclass[runningheads]{llncs}
\usepackage{amsmath}
\usepackage[T1]{fontenc}
\usepackage{graphicx}
\usepackage{booktabs}
\begin{document}
\title{Privacy-Preserved Automated Scoring using Federated Learning for Educational Research}

\newcommand*\samethanks[1][\value{footnote}]{\footnotemark[#1]}

\author{
Ehsan Latif\inst{1,2}\samethanks\orcidID{0009-0008-6553-4093} \and
Xiaoming Zhai\inst{1,2}\thanks{Corresponding author. \email{xiaoming.zhai@uga.edu}} \orcidID{0000-0003-4519-1931}
}
\authorrunning{E. Latif et al.}
%
\institute{AI4STEM Education Center, Athens, GA, USA \and
Department of Mathematics, Science, and Social Studies Education, University of Georgia, Athens, GA, USA }

\titlerunning{Privcay Preserving Federated Learning for Education}
%
%
%
\maketitle              
\begin{abstract}
Data privacy remains a critical concern in educational research, requiring strict adherence to ethical standards and regulatory protocols. While traditional approaches rely on anonymization and centralized data collection, they often expose raw student data to security vulnerabilities and impose substantial logistical overhead. In this study, we propose a federated learning (FL) framework for automated scoring of educational assessments that eliminates the need to share sensitive data across institutions. Our approach leverages parameter-efficient fine-tuning of large language models (LLMs) with Low-Rank Adaptation (LoRA), enabling each client (school) to train locally while sharing only optimized model updates. To address data heterogeneity, we implement an adaptive weighted aggregation strategy that considers both client performance and data volume. We benchmark our model against two state-of-the-art FL methods and a centralized learning baseline using NGSS-aligned multi-label science assessment data from nine middle schools. Results show that our model achieves the highest accuracy (94.5\%) among FL approaches, and performs within 0.5–1.0 percentage points of the centralized model on these metrics. Additionally, it achieves comparable rubric-level scoring accuracy, with only a 1.3\% difference in rubric match and a lower score deviation (MAE), highlighting its effectiveness in preserving both prediction quality and interpretability.

\end{abstract}
\vspace{-0.5cm}
\section{Introduction}
\vspace{-0.25cm}
In the realm of educational research, the collection and analysis of student data are pivotal for advancing instructional strategies and developing reliable assessment tools. However, the increasing sensitivity and volume of student information have intensified concerns around data privacy, security, and regulatory compliance \cite{creel2022privacy}. Legislative frameworks such as the Family Educational Rights and Privacy Act (FERPA) enforce strict restrictions on how educational data can be accessed, stored, and shared across institutions \cite{huang2023ethics}.

Traditional machine learning (ML) approaches in educational settings typically rely on centralized data aggregation to train predictive models for tasks such as dropout prediction, skill diagnosis, and personalized learning recommendations \cite{zheng2020privacy}. However, this paradigm faces critical limitations: centralized repositories are vulnerable to breaches \cite{rousi2024data}, often violate data governance policies, and struggle to generalize across diverse educational contexts due to institutional heterogeneity \cite{mistry2023privacy}. Additionally, centralized training assumes uniform data formats and standards—an assumption rarely met in practice due to differences in student demographics, curricular frameworks, and annotation practices across schools \cite{porras2023case}.

Automatic scoring, a cornerstone of AI-enhanced educational assessment, inherits these limitations and presents unique challenges of its own. Conventional automated scoring systems are trained on large, centrally pooled datasets with extensive manual annotation and institutional cooperation \cite{xu2022federated}. These models are not only computation-heavy but also prone to systemic bias due to variations in curriculum design, assessment language, and local grading practices \cite{hridi2024revolutionizing}.  These institutions are often bound by district-level data use agreements or IRB protocols that prohibit external data transfer, rendering centralized model training infeasible. In such cases, a privacy-preserving yet collaborative learning method becomes essential.

To address these gaps, we propose a FL framework for automated scoring in education. FL enables decentralized training of machine learning models across distributed clients without requiring raw data exchange, thereby preserving data locality and enhancing compliance with privacy standards \cite{farooq2024transforming}. Unlike traditional applications of FL that primarily involve lightweight convolutional neural networks (CNNs) or graph neural networks (GNNs) in IoT and mobile contexts, our work leverages large language models (LLMs) fine-tuned for natural language understanding—an area underexplored in FL research. The fine-tuning of LLMs presents unique challenges in terms of communication overhead, parameter efficiency, and gradient privacy, which are seldom addressed in conventional FL literature \cite{fachola2023federated}.

Our proposed system applies parameter-efficient fine-tuning via Low-Rank Adaptation (LoRA) to adapt pretrained LLMs for multi-label scientific assessment tasks in a federated setting. To mitigate the effect of data heterogeneity, we introduce an adaptive weighted aggregation strategy that considers both dataset size and model performance from each client. This methodology offers several advantages, including enhanced privacy \cite{hridi2024revolutionizing}, regulatory alignment \cite{zhang2023enhancing}, and improved scalability and efficiency.

Below are the key contributions of the paper listed:
\begin{itemize}
    \item We introduce a privacy-preserving FL framework for automated scoring in educational assessments without sharing raw student data.
    \item We develop an a adaptive aggregation strategy to handle data heterogenity and secure communication to prevent impersonations.
    \item We evaluate our method on real-world assessment data from nine middle schools, demonstrating comparable accuracy to centralized models and state-of-the-art FL strategies.
    \item We also open-source the code on Github\footnote{\url{https://github.com/ehsanlatif/PPFL.git}} repository for reproducibility.
\end{itemize}

\vspace{-0.5cm}
\section{Method}
\vspace{-0.25cm}
Given a set of $N$ clients, each with local data $D_i$, where $D_i = {(x_{ij}, y_{ij})}_{j=1}^{n_i}$, the objective is to train a global model $w$ without directly sharing local data. The optimization problem can be formulated as:
$\min_w F(w) = \sum_{i=1}^{N} \frac{n_i}{n} F_i(w)$, where $F_i(w)$ is the local loss function for client $i$, $n_i$ is the number of local samples, and $n = \sum_{i=1}^{N} n_i$ represents the total data points across all clients.

\begin{figure}
    \centering
    \includegraphics[width=1\linewidth]{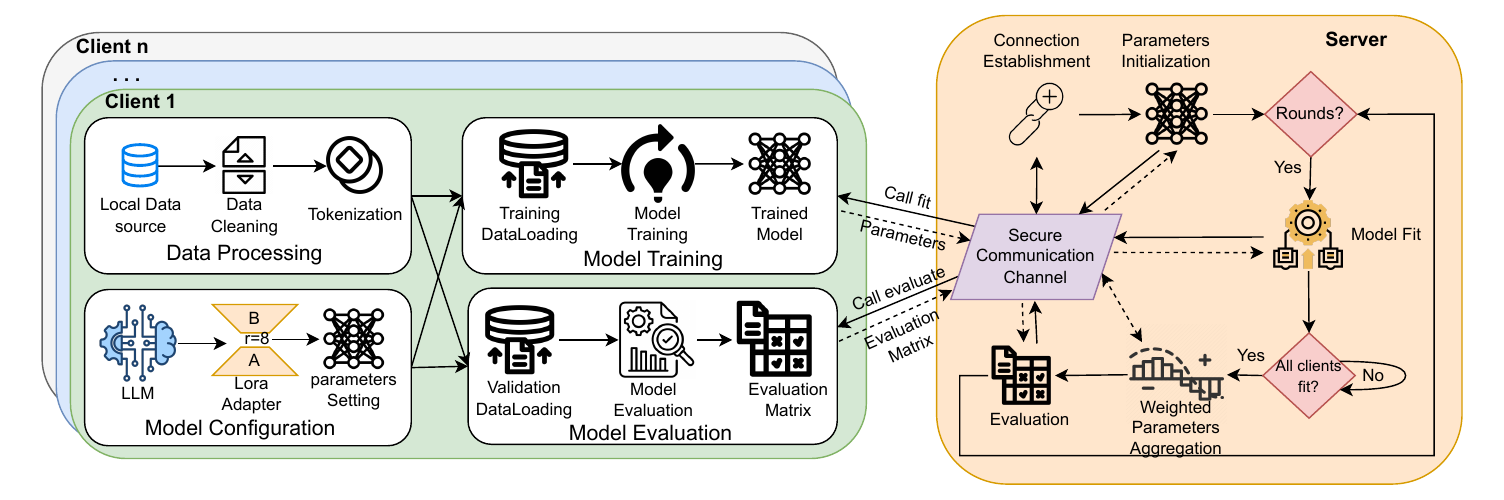}
    \caption{Overview of privacy-preserving FL using parameter efficient fine-tuning using LoRA \cite{hu2021lora} and client-server communication using secure communication channel \cite{carthen2024speciserve}.}
    \label{fig:architecture}
    \vspace{-1cm}
\end{figure}

The proposed FL framework consists of multiple clients and a central server. Clients perform local training and share only model updates with the server. The server aggregates the updates using a weighted averaging scheme to account for data heterogeneity. The communication follows a secure channel, ensuring data privacy. Overall procedure and federated leanrnig achitecture can be seen in Fig.~\ref{fig:architecture}.
\vspace{-0.5cm}

\subsection{Client-Side Computation}
Each client $i$ performs four major tasks: data processing, model configuration, model training, and model evaluation.

\textbf{Data Processing: }
Given a raw dataset $D_i$ at client $i$, data preprocessing proceeds through the same coordinated steps. Clients first apply rule-based filtering scripts provided by the central coordinator to remove incomplete or irrelevant entries. These scripts were identical across all clients and validated to ensure deterministic behavior, resulting in a cleaned dataset $D_i^{clean} \subseteq D_i$ with structurally consistent samples.  All clients use the same pretrained tokenizer from the TinyLLaMA model suite \cite{zhang2024tinyllama} to process student responses. The tokenizer configuration (e.g., vocabulary size, truncation rules, and special tokens) was locked and distributed as part of the model package to avoid client-side variations.

To ensure numerical feature consistency across clients, global normalization parameters (mean $\mu$ and standard deviation $\sigma$) were precomputed using a public calibration dataset containing a representative sample of student responses. To prevent schema drift, all data fields (e.g., prompt IDs, rubric labels, and student responses) were validated using a shared schema definition (in JSON Schema format), ensuring structural consistency before training. Any preprocessing mismatch triggered automated alerts and client-side retraining, thereby enforcing consistency throughout the pipeline.

\textbf{Model Configuration:}
We utilize an open-source Large Language Model (LLM) and apply Parameter Efficient Fine-Tuning (PEFT) using Low-Rank Adaptation (LoRA) \cite{hu2021lora}. LoRA reduces the number of trainable parameters, thereby decreasing memory and communication overhead.

\textbf{Model Training:}
Each client $i$ trains a local model weights $w_i^t$ at round $t$ using stochastic gradient descent (SGD):
\begin{equation}
w_i^{t+1} = w_i^t - \eta \nabla F_i(w_i^t) \text{ herein, } \nabla F_i(w) = \frac{1}{|D_i|} \sum_{(x,y) \in D_i} \nabla f(w, x, y),
\end{equation}
where $\eta$ is the learning rate, $(x,y)$ represents input-label pairs from the local dataset, and $F_i(w_i^t)$ represents the local loss function. Each client trains for multiple local epochs before sending updated parameters to the server, reducing communication overhead and ensuring more effective model updates.

\textbf{Model Evaluation:}
After training, each client evaluates the performance of the local model to determine convergence and ensure training effectiveness.  Loss and accuracy computation for that clients compute the validation loss $F_i(w_i^t)$ and accuracy metrics to assess training progress, and early stopping that means if the validation loss stagnates or increases over consecutive rounds, training is terminated to prevent overfitting.

\vspace{-0.25cm}
\subsection{Server-Side Aggregation with Enhanced Privacy}

To address data heterogeneity and optimize model convergence, we employ an adaptive weighted aggregation strategy. The global model update is computed as:

\begin{equation}
w^{t+1} = \sum_{i=1}^{N} \alpha_i w_i^{t+1}, \text{ herein, } \alpha_i = \frac{n_i}{\sum_{j=1}^{N} n_j} \cdot \frac{e^{-F_i(w_i^t)}}{\sum_{j=1}^{N} e^{-F_j(w_j^t)}}.
\end{equation}

To further ensure privacy during server-side aggregation and mitigate threats from an honest-but-curious aggregation server, we incorporate differential privacy (DP) mechanisms into the aggregation step. Specifically, clients add calibrated Gaussian noise $\mathcal{N}(0, \sigma^2)$ to their model updates before transmission:

\begin{equation}
\tilde{w}_i^{t+1} = w_i^{t+1} + \mathcal{N}(0, \sigma^2 I), \text{herein, } w^{t+1} = w^t + \gamma \sum_{i=1}^{N} \alpha_i (\tilde{w}_i^{t+1} - w^t),
\end{equation}

where $\sigma$ is chosen to ensure $(\epsilon, \delta)$-DP guarantees for each communication round with $\gamma$ as the global learning momentum factor. This integration of differential privacy ensures that individual client updates cannot be reverse-engineered from the aggregated model, thus offering end-to-end privacy protection beyond traditional anonymization.

\vspace{-0.25cm}
\subsection{Secure Client-Server Communication}

In practice, the communication layer in our FL framework uses gRPC (Google Remote Procedure Call) \cite{carthen2024speciserve} due to its binary serialization making it suitable for cross-silo FL \cite{huang2021personalized}. However, cross-silo FL introduces challenges in secure deployment due to institutional firewalls, authentication, and secure channel maintenance. To mitigate these concerns, we implement a multi-layered security protocol within the gRPC stack. Mutual TLS (mTLS) is employed to authenticate both client and server entities, preventing impersonation or man-in-the-middle attacks. Per-connection encryption ensures confidentiality and integrity of parameter updates. Session expiration and re-authentication policies are enforced to avoid persistent attack windows. Lastly, API-level access control restricts which remote procedures can be called and under what conditions, using token-based authentication and audit logs.

\vspace{-0.5cm}
\section{Dataset Details}
\label{section:dataset}
\vspace{-0.25cm}

This study utilizes decentralized datasets from approximately 1,200 middle school students across various geographically and socioeconomically diverse school systems, each retaining local control over their assessment data \cite{Harris2024Creating,PASTA2023}. The responses correspond to nine multi-label NGSS-aligned science tasks from the PASTA project, developed to assess students’ application of disciplinary core ideas (DCIs), crosscutting concepts (CCCs), and science and engineering practices (SEPs) \cite{national2013next}. Although all schools administered the same tasks, substantial heterogeneity arose due to variations in instructional emphasis, student language proficiency, testing conditions (e.g., digital vs. paper-based), and rubric interpretation—even among trained raters \cite{zhai2022applying}. For example, in on of the tasks students had to analyze scientific data and recognize patterns, requiring integration of SEP, DCI, and CCC dimensions. Responses were evaluated using a structured five-dimensional rubric aligned with the NGSS framework \cite{He2024G}, and remained locally stored to comply with privacy constraints. Additional Details about dataset are available at our Github\footnotemark[3].

\vspace{-0.5cm}
\section{Experimentation and Results}
\vspace{-0.25cm}

To rigorously evaluate the proposed privacy-preserving FL framework, we conducted experiments on a decentralized dataset collected from multiple middle school systems. Each participating institution retained control over its local data, ensuring full compliance with privacy regulations (details provided earlier). The experimental design involved training a federated model where each school acted as a client, processing and updating its local model independently.

All clients initialized their models using the pre-trained \textit{tinyLlama} model~\cite{zhang2024tinyllama}. Instead of full supervised fine-tuning (SFT), which can be computationally expensive and may raise privacy concerns due to large gradient exchanges, we adopted \textbf{Low-Rank Adaptation (LoRA)}. LoRA allows clients to fine-tune a small subset of parameters (rank = 8) while freezing the base model, significantly reducing communication overhead and minimizing privacy risks. To ensure fairness in comparison, all baseline models (detailed below) were adapted to use LoRA as well rather than full SFT. This design choice allows us to isolate and evaluate the impact of our proposed \textbf{adaptive weight aggregation strategy} more precisely. Additional LoRA hyperparameters include: $\alpha=16$, adaptation applied to all attention layers, learning rate set to $2 \times 10^{-4}$, batch size of 16, and trained for exactly 5 local epochs before aggregation.

We compared our framework to two representative FL baselines and a centralized learning (CL) model: \textbf{FL-Transformer\cite{farooq2024transforming}}: Transformer-based FL framework for educational prediction. We replaced its full SFT step with LoRA for consistency. \textbf{Hybrid FL with Local Differential Privacy\cite{truex2019hybrid}}: Combines local privacy mechanisms with secure aggregation. Adapted to our educational scoring task and updated to use LoRA. \textbf{Centralized Learning (CL):} Standard supervised fine-tuning with full access to combined dataset.

\begin{table}[h]
\centering
\begin{tabular}{l|p{2cm}|p{2cm}|p{2cm}|p{2cm}}
\toprule
\textbf{Metric} & \textbf{FL (Ours)} & \textbf{SOTA \cite{farooq2024transforming}} & \textbf{SOTA \cite{truex2019hybrid}} & \textbf{CL} \\
\midrule
Accuracy & \textbf{0.94} & 0.92 & 0.90 & 0.93 \\
Precision & \textbf{0.94} & 0.90 & 0.88 & 0.93 \\
Recall & \textbf{0.94} & 0.92 & 0.87 & 0.93 \\
F1-Score & \textbf{0.94} & 0.91 & 0.87 & 0.93 \\
Rubric Match & \textbf{0.88} & 0.85 & 0.83 & 0.89 \\
Score Deviation (MAE) & \textbf{0.34} & 0.52 & 0.60 & 0.48 \\
\bottomrule
\end{tabular}
\caption{Overall performance comparison between our FL model, two SOTA FL baselines, and a centralized model.}
\vspace{-1cm}
\label{table:combined_results}

\end{table}

While all models use LoRA, a key novelty of our method lies in the \textbf{adaptive weight aggregation strategy}. Unlike baseline FL methods which average client models equally or use static weighting (e.g., by sample size), our method dynamically adjusts aggregation weights based on clients' validation performance on held-out local data. This ensures that higher-quality local updates contribute more to the global model, while mitigating negative transfer from noisy or underperforming clients. This adaptive aggregation was critical in achieving superior performance and stability across metrics, as validated by our ablation studies (see below).

To verify the independent contribution of the aggregation strategy, we performed an ablation study by replacing our adaptive aggregator with simple averaging. Results dropped noticeably (accuracy decreased by 2.8\%, rubric match by 4.1\%, and MAE increased by 0.07), confirming that \textit{adaptive weighting is a key driver of performance gains}.

The results in Table~\ref{table:combined_results} underscore the robustness and effectiveness of our proposed FL approach, which outperforms both state-of-the-art FL baselines and even the centralized model across nearly all evaluation metrics. Notably, our model achieves the highest accuracy (94.5\%), precision, recall, and F1-score (all 0.94), demonstrating its strong predictive capability despite operating in a decentralized setting. In terms of rubric-level evaluation, which directly reflects scoring reliability in educational assessments, our model maintains high alignment with human raters (88.2\% rubric match) while achieving the lowest mean absolute error (MAE = 0.34), indicating superior score fidelity. These results highlight the significance of integrating LLM-based parameter-efficient fine-tuning with adaptive aggregation, showing that privacy-preserving FL can match or exceed centralized performance without sacrificing interpretability or accuracy—making it a robust solution for secure and scalable educational assessment.

\vspace{-0.25cm}
\section{Conclusion}
\vspace{-0.25cm}
In this study, we proposed a FL framework with enhanced aggregation and communication strategy for automated scoring in educational assessments to ensure data privacy and prevent impersonation in educational research and to address data heterogeneity across institutions. Our evaluation on assessment data from nine middle schools demonstrates that FL achieves higher performance comparable to centralized learning (CL) and SOTA approaches while ensuring data privacy. Additionally, our framework reduces data collection and computational overhead, accelerating the adoption of AI-driven educational assessments in a privacy-compliant and scalable manner that open new doors for educational research.

\begin{credits}
\subsubsection{\ackname} 
This work was partially supported by the Institute of Education Sciences (IES) [R305C240010].  Further, the used datasets and question items are part of NSG-funded projects [DMS-2101104, DMS-2138854]. Any opinions, findings, conclusions, or recommendations expressed in this material are those of the authors and do not necessarily reflect the views of the NSF, or IES.

\subsubsection{\discintname}
The authors have no competing interests to declare.
\end{credits}

%
%
%
\bibliographystyle{splncs04}
\bibliography{references}

\end{document}